\documentclass[11pt,a4paper]{article}
\usepackage[hyperref]{emnlp2018}
\usepackage{times}
\usepackage{latexsym}
\usepackage{amsmath}
\usepackage{multirow}
\usepackage{url}
\usepackage{subcaption}
\makeatletter

\newcommand{\@emptybiblabel}[1]{}

\makeatother

\usepackage{algorithmicx, algorithm, algpseudocode}

\usepackage[utf8]{inputenc}
\usepackage{amsmath}
\usepackage{amssymb}
\usepackage{wasysym}
\usepackage{setspace}
\usepackage{color}
\usepackage{booktabs}
\usepackage{microtype}
\usepackage{tikz-dependency}
\usepackage{hhline}
\usepackage[colorinlistoftodos]{todonotes}
\usepackage{titlesec}

\newcommand{\smallpar}[1]{\vspace{-3pt}\paragraph{#1}}
\newcommand{\smallsection}[1]{\vspace{-5pt}\section{#1}}
\newcommand{\smallsubsection}[1]{\vspace{-6pt}\subsection{#1}}

\usepackage{blindtext}
\algrenewcommand\alglinenumber[1]{{\sffamily\footnotesize#1}}

\newcommand{\eat}[1]{\ignorespaces}

\usepackage{color}

\newenvironment{nospaceflalign*}
 {\setlength{\abovedisplayskip}{0pt}\setlength{\belowdisplayskip}{0pt}%
  \csname flalign*\endcsname}
 {\csname endflalign*\endcsname\ignorespacesafterend}

\eat{
\setlength{\abovedisplayskip}{2pt}
\setlength{\belowdisplayskip}{2pt}

\titlespacing{\section}{0pt}{-0.5\parskip}{-0.6\parskip}
\titlespacing{\subsection}{0pt}{-0.5\parskip}{-0.5\parskip}
\titlespacing{\subsubsection}{0pt}{-0.5\parskip}{-0.5\parskip}

\captionsetup[table]{belowskip=0pt}
}

\newcommand{\specificthanks}[1]{\@fnsymbol{#1}}

\aclfinalcopy % Uncomment this line for the final submission
\title{Deep Probabilistic Logic: A Unifying Framework for Indirect Supervision}

\author{Hai Wang\textsuperscript{1}\thanks{~~This work was conducted at Microsoft Research.} ~~ Hoifung Poon\textsuperscript{2}\\
\textsuperscript{1} Toyota Technological Institute at Chicago, Chicago, Illinois, USA \\ \textsuperscript{2} Microsoft Research, Redmond, WA, USA \\
{\tt haiwang@ttic.edu ~~~ hoifung@microsoft.com}
}

\hypersetup{draft}

\begin{document}

\maketitle

\begin{abstract}

Deep learning has emerged as a versatile tool for a wide range of NLP tasks, due to its superior capacity in representation learning. But its applicability is limited by the reliance on annotated examples, which are difficult to produce at scale. Indirect supervision has emerged as a promising direction to address this bottleneck, either by introducing labeling functions to automatically generate noisy examples from unlabeled text, or by imposing constraints over interdependent label decisions. A plethora of methods have been proposed, each with respective strengths and limitations. Probabilistic logic offers a unifying language to represent indirect supervision, but end-to-end modeling with probabilistic logic is often infeasible due to intractable inference and learning. In this paper, we propose deep probabilistic logic (DPL) as a general framework for indirect supervision, by composing probabilistic logic with deep learning. DPL models label decisions as latent variables, represents prior knowledge on their relations using weighted first-order logical formulas, and alternates between learning a deep neural network for the end task and refining uncertain formula weights for indirect supervision, using variational EM. This framework subsumes prior indirect supervision methods as special cases, and enables novel combination via infusion of rich domain and linguistic knowledge. Experiments on biomedical machine reading demonstrate the promise of this approach.

\end{abstract}

%\footnote{We will release code upon publication.}

\begin{figure}
    \centering
    \includegraphics[width=0.9\linewidth]{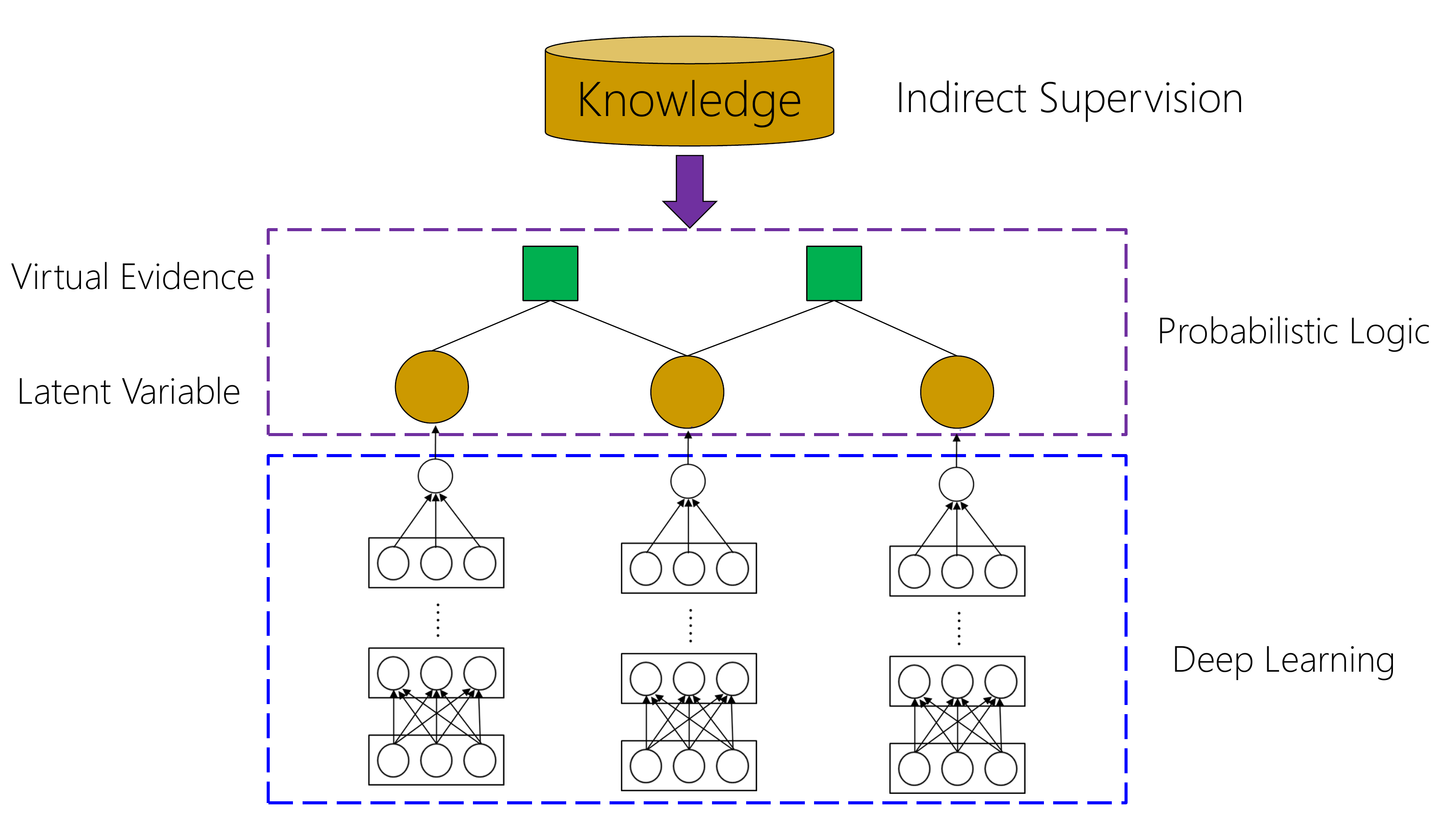}
    \vspace{-5pt}
    \caption{Deep Probabilistic Logic: A general framework for combining indirect supervision strategies by composing probabilistic logic with deep learning. Learning amounts to maximizing conditional likelihood of virtual evidence given input by summing up latent label decisions.
    }
    \label{fig:DPL}
\end{figure}

\smallsection{Introduction}

Deep learning has proven successful in a wide range of NLP tasks \cite{bahdanau2014neural,bengio2003neural,clark2016improving,hermann2015teaching,sutskever2014sequence}.
%Its versatility stems from its capacity to compactly represent complex functions and word embedding that can be trained from large amount of available text \cite{bengio or DL book,word2vec,glove}.
The versatility stems from its capacity of learning a compact representation of complex input patterns \cite{goodfellow2016deep}. %,mikolov2013distributed,pennington2014glove}.
However, success of deep learning is bounded by its reliance on labeled examples, which are expensive and time-consuming to produce. 
%This challenge is particularly pronounced in high-value domains like biomedicine, where crowd-sourcing is difficult to apply.
Indirect supervision has emerged as a promising direction for breaching the annotation bottleneck.
%, and a variety of methods have been proposed to leverage more readily available resources to compensate for the lack of labeled examples.
A powerful paradigm is {\em joint inference} \cite{chang&al07,poon&domingos08,druck&mccallum08,ganchev&al10}, which leverages linguistic and domain knowledge to impose constraints over interdependent label decisions. 
More recently, another powerful paradigm, often loosely called {\em weak supervision}, has gained in popularity. The key idea is to introduce labeling functions to automatically generate (noisy) training examples from unlabeled text.
{\em Distant supervision} is a prominent example that used existing knowledge bases for this purpose \cite{craven1999constructing,mintz2009distant}. {\em Data programming} went further by soliciting labeling functions from domain experts \cite{ratner&al16,bach&al17}. 

Indirect-supervision methods have achieved remarkable successes in a number of NLP tasks, but they also exhibit serious limitations.
%Like bootstrap learning \cite{?}, d
Distant supervision often produces incorrect labels, whereas labeling functions from data programming vary in quality and coverage, and may contradict with each other on individual instances.
Joint inference incurs greater modeling complexity and often requires specialized learning and inference procedures.

Since these methods draw on diverse and often orthogonal sources of indirect supervision, combining them may help address their limitations and amplify their strengths.
Probabilistic logic offers an expressive language for such an integration, and is well suited for resolving noisy and contradictory information \cite{richardson&domingos06}.
Unfortunately, probabilistic logic generally incurs intractable learning and inference, often rendering end-to-end modeling infeasible.

In this paper, we propose {\bf deep probabilistic logic (DPL)} as a unifying framework for indirect supervision (Figure~\ref{fig:DPL}). %, drawing inspiration from recent work using logical rules to aid deep learning \cite{hu2016harnessing,hu2016deep}. 
Specifically, we made four contributions.
First, we introduce a modular design to compose probabilistic logic with deep learning, with a supervision module that represents indirect supervision using probabilistic logic, and a prediction module that performs the end task using a deep neural network.
Label decisions are modeled as latent variables and serve as the interface between the two modules.

Second, we show that all popular forms of indirect supervision can be represented in DPL by generalizing virtual evidence \cite{subramanya&bilmes07, pearl2014probabilistic}. %to arbitrary potential functions representing hard and soft constraints. 
Consequently, these diverse methods can be easily combined within a single framework for mutual amplification.

Third, we show that our problem formulation yields a well-defined learning objective (maximizing conditional likelihood of virtual evidence).
%, which opens up a variety of optimization strategies. %objective is end-to-end differentiable and can be directly optimized. 
We proposed a modular learning approach by decomposing the optimization over the supervision and prediction modules, using variational EM, which enables us to apply state-of-the-art methods for probabilistic logic and deep learning.

\begin{figure}
    \centering
    \includegraphics[width=0.9\linewidth]{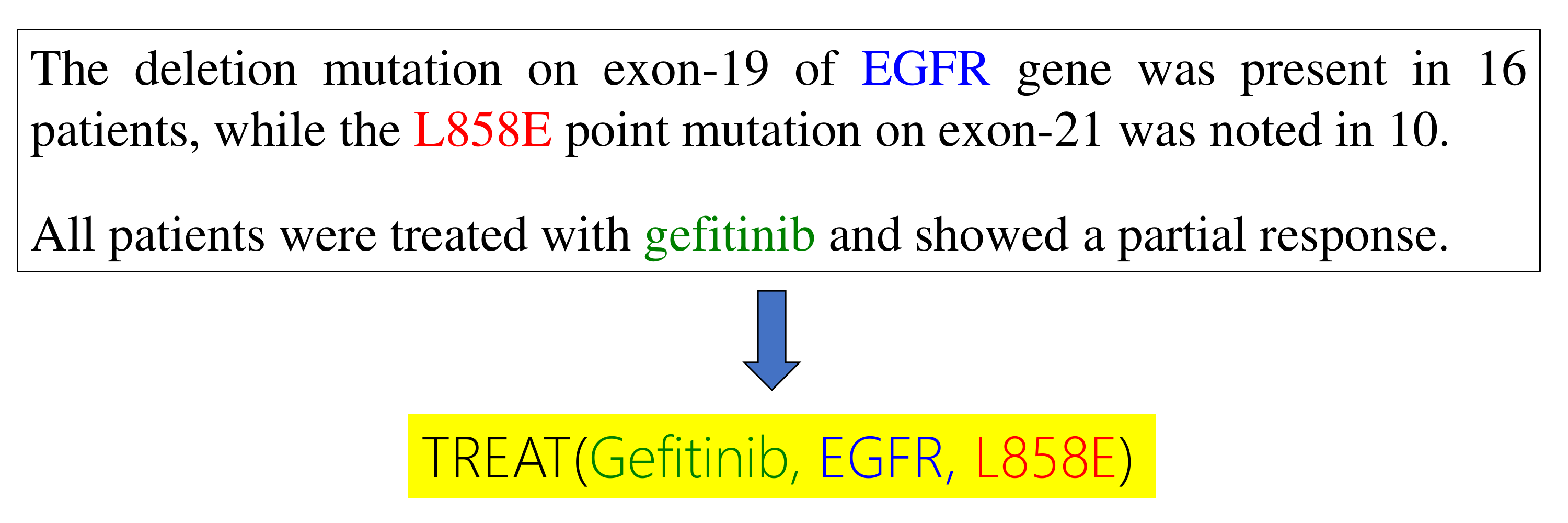}
    \vspace{-5pt}
    \caption{Example of cross-sentence relation extraction for precision cancer treatment.
    }
    \label{fig:relextract-example}
\end{figure}

Finally, we applied DPL to biomedical machine reading \cite{quirkpoon2017,peng&al17}.
Biomedicine offers a particularly attractive application domain for exploring indirect supervision. 
%The advent of big biomedical data such as cancer genome sequences heralds the new era of precision medicine \cite{nature}, but interpreting such data requires assimilating a vast amount of published knowledge from text, making it a priority to develop machine reading methods for automating knowledge curation   
Biomedical literature grows by over one million each year\footnote{\url{http://ncbi.nlm.nih.gov/pubmed}}, making it imperative to develop machine reading methods for automating knowledge curation (Figure~\ref{fig:relextract-example}).
While crowd sourcing is hardly applicable, there are rich domain knowledge and structured resources to exploit for indirect supervision.
%In this paper, we use entity linking and cross-sentence relation extraction as our example tasks to demonstrate how a variety of weak supervision and denoising strategies can be combined in the KRGM framework to improve predictive accuracy.
%, which is relatively underexplored compared to newswire and web domains, and presents unique challenges for weak supervision.
Using cross-sentence relation extraction and entity linking as case studies, we show that distant supervision, data programming, and joint inference can be seamlessly combined in DPL to substantially improve machine reading accuracy, without requiring any manually labeled examples.\footnote{The DPL code and datasets will be made available at \url{ http://hanover.azurewebsites.net}.}

\smallsection{Related Work}

\smallpar{Distant supervision}
This paradigm was first introduced for binary relation extraction \cite{craven1999constructing,mintz2009distant}. 
In its simplest form, distant supervision generates a positive example if an entity pair with a known relation co-occurs in a sentence, and samples negative examples from co-occurring entity pairs not known to have the given relation.
%leverages available databases to automatically annotate training examples in unlabeled text. It 
It has recently been extended to cross-sentence relation extraction \cite{quirkpoon2017,peng&al17}. In principle, one simply looks beyond single sentences for co-occurring entity pairs. However, this can introduce many false positives and prior work used a small sliding window and filtering (minimal-span) to mitigate training noise. Even so, accuracy is relatively low. 
Both \newcite{quirkpoon2017} and \newcite{peng&al17} used ontology-based string matching for entity linking, which also incurs many false positives, as biomedical entities are highly ambiguous (e.g., PDF and AAAS are gene names). %, and ER can refer to endrogen receptor (a gene) or emergency room).
%They used a set of filtering heuristics to reduce precision errors, at the expense of significant recall loss.
%This introduced many errors, and both observed that a significant portion of extraction mistakes (?\%-?\%) stem from entity linking errors, where entities are mistaken as genes or drugs.
Distant supervision for entity linking is relatively underexplored, and prior work generally focuses on Freebase entities, where links to the corresponding Wikipedia articles are available for learning \cite{huang2015leveraging}.
%Such weak supervision is generally not available for entities in specialized domains like biomedicine.
%{\bf Move some to EL?}
%Instead, we will explore a distant supervision approach that only assumes that a lexicon exists for the entities, which is usually readily available in domain ontologies. (e.g., HUGO for human genes, DrugBank for drugs, ...).
%Mention biomed: NERD - focus on NER.

\smallpar{Data Programming} 
Instead of annotated examples, domain experts are asked to produce labeling functions, each of which assigns a label to an instance if the input satisfies certain conditions, often specified by simple rules \cite{ratner&al16}. 
%While this may not apply to domains with complex input patterns (e.g., images or gene expression profiles), 
This paradigm is useful for semantic tasks, as high-precision text-based rules are often easy to come by. 
%(\textcolor{red}{slot filling in dialog manager})
However, there is no guarantee on broad coverage, and labeling functions are still noisy and may contradict with each other.
The common denoising strategy assumes that labeling functions make random mistakes, and focuses on estimating their accuracy and correlation \cite{ratner&al16,bach&al17}.
A more sophisticated strategy also models instance-level labels and uses instance embedding to estimate instance-level weight for each labeling function \cite{liu2017heterogeneous}.

\smallpar{Joint Inference}

%Distant supervision and data programming focus on weak supervision concerning individual labels. 
Distant supervision and data programming focus on infusing weak supervision on individual labels. 
Additionally, there is rich linguistic and domain knowledge that does not specify values for individual labels, but imposes hard or soft constraints on their joint distribution.
For example, if two mentions are coreferent, they should agree on entity properties \cite{poon&domingos08}.
There is a rich literature on joint inference for NLP applications. %\cite{mccallum09}. %
Notable methodologies include constraint-driven learning \cite{chang&al07}, general expectation \cite{druck&mccallum08}, posterior regularization \cite{ganchev&al10}, and probabilistic logic \cite{poon&domingos08}. %\cite{poon&domingos07,poon&vanderwende10}.
Constraints can be imposed on relational instances or on model expectations.
Learning and inference are often tailor-made for each approach, including beam search, primal-dual optimization, weighted satisfiability solvers, etc.
Recently, joint inference has also been used in denoising distant supervision. 
Instead of labeling all co-occurrences of an entity pair with a known relation as positive examples, one only assumes that at least one instance is positive \cite{MultiR,lin&al16}. 
%This is either enforced as a hard constraint \cite{MultiR} or as a weighted sum using attention mechanism \cite{lin&al16}. 

\smallpar{Probabilistic Logic}

Probabilistic logic combines logic's expressive power with graphical model's capability in handling uncertainty. A representative example is Markov logic \cite{richardson&domingos06}, which define a probability distribution using weighted first-order logical formulas as templates for a Markov model.
%being a . Given random variables $X$ for a problem domain,  Markov logic uses a set of weighted first-order logical formulas $(w_i, f_i): i$ to define a probability distribution $P(X)\propto \exp\sum_i~w_i\cdot f_i(X)$. 
Probabilistic logic has been applied to incorporating indirect supervision for various NLP tasks \cite{poon&domingos07,poon&domingos08,poon&vanderwende10}, but its expressive power comes at a price: learning and inference are generally intractable, and end-to-end modeling often requires heavy approximation 
%, such as relaxation to linear programming 
\cite{kimmig2012short}.
In DPL, we limit the use of probabilistic logic to modeling indirect supervision in the supervision module, leaving end-to-end modeling to deep neural network in the prediction module.
This alleviates the computational challenges in probabilistic logic, while leveraging the strength of deep learning in distilling complex patterns from high-dimension data.

\smallpar{Knowledge-Rich Deep Learning}
%\smallpar{Knowledge-rich deep learning}

%Graphical models are expressive and easy to interpret, yet inference is challenging for high-dimension data with complex dependencies.

Infusing knowledge in neural network training is a long-standing challenge in deep learning \cite{towell1994knowledge}.
\newcite{hu2016harnessing,hu2016deep} first used logical rules to help train a convolutional neural network for sentiment analysis. 
DPL draws inspiration from their approach, but is more general and theoretically well-founded.
\newcite{hu2016harnessing,hu2016deep} focused on supervised learning and the logical rules were introduced to augment labeled examples via posterior regularization \cite{ganchev&al10}. 
DPL can incorporate both direct and indirect supervision, including posterior regularization and other forms of indirect supervision.
Like DPL, \newcite{hu2016deep} also refined uncertain weights of logical rules, but they did it in a heuristic way by appealing to symmetry with standard posterior regularization. We provide a novel problem formulation using generalized virtual evidence, which shows that their heuristics is a special case of variational EM and opens up opportunities for other optimization strategies.

%yields a well-defined learning objective (Section~\ref{sec:mdl}). This opens up opportunities for various optimization strategies and shows that their heuristic approach is a special case of variational EM.

Deep generative models also combine deep learning with probabilistic models, but focus on uncovering latent factors to support generative modeling and semi-supervised learning \cite{kingma&welling13,kingma&al14}.
Knowledge infusion is limited to introducing structures among the latent variables (e.g., Markov chain) %for better modeling observed data
\cite{johnson&al16}.
%{\em Deep probabilistic programming} provides a flexible interface for exploring such composition \cite{edward}.
In DPL, we focus on learning a discriminative model for predicting the latent labels, using a probabilistic model defined by probabilistic logic to inject indirect supervision.

\smallsection{Deep Probabilistic Logic}
\label{sec:mdl}

In this section, we introduce deep probabilistic logic (DPL) as a unifying framework for indirect supervision.
%The key idea is to model label decisions as latent variables, and introduce a  supervision module using Markov logic, which defines a probabilistic distribution over the latent label variables.
Label decisions are modeled as latent variables. Indirect supervision is represented as generalized virtual evidence, %, which are potential functions on label variables and inputs.
and learning maximizes the conditional likelihood of virtual evidence given input. %, which can be done by variational EM that alternates between estimating marginal probability of labels (E-step), as well as using these probabilistic labels to train the deep neural network in the prediction module and refine uncertain parameters in the supervision module (M-step).
We first review the idea of virtual evidence and show how it can be generalized to represent any form of indirect supervision. % as soft and hard constraints over labels and inputs.
We then formulate the learning objective and show how it can be optimized using variational EM.

Given a prediction task, let $\mathcal{X}$ denote the set of possible inputs and $\mathcal{Y}$ the set of possible outputs. The goal is to train a prediction module $\Psi(x,y)$ that scores output $y$ given input $x$. 
%that estimates the conditional probability of output $y$ given input $x$.
Without loss of generality, we assume that $\Psi(x,y)$ defines the conditional probability $P(y|x)$ using a deep neural network with a softmax layer at the top.
Let $X=(X_1,\cdots,X_N)$ denote a sequence of inputs and $Y=(Y_1,\cdots,Y_N)$ the corresponding outputs. We consider the setting where $Y$ are unobserved, and $\Psi(x,y)$ is learned using indirect supervision.

\smallpar{Virtual evidence} 
Pearl \cite{pearl2014probabilistic} first introduced the notion of virtual evidence, which has been used to incorporate label preference in semi-supervised learning \cite{reynolds&bilmes05,subramanya&bilmes07,xiao09} and grounded learning \cite{parikh&al15}.
%to represent prior belief on the value of a random variable in a Bayesian network. 
%To illustrate the idea, let's consider a simple case.
Suppose we have a prior belief on the value of $y$, it can be represented by introducing a binary variable $v$ as a dependent of $y$ such that $P(v=1|y=l)$ is proportional to the prior belief of $y=l$.
$v=1$ is thus an observed evidence that imposes soft constraints over $y$. 
Direct supervision (i.e., observed label) for $y$ is a special case when the belief is concentrated on a specific value $y=l^*$ (i.e., $P(v=1|y=l)=0$ for any $l\ne l^*$). 
The virtual evidence $v$ can be viewed as a reified variable for a potential function $\Phi(y)\propto P(v=1|y)$. 
This enables us to generalize virtual evidence to arbitrary potential functions $\Phi(X,Y)$ over the inputs and outputs.
In the rest of the paper, we will simply refer to the potential functions as virtual evidences, without introducing the reified variables explicitly.

\smallpar{DPL}
Let $K=(\Phi_1,\cdots,\Phi_V)$ be a set of virtual evidence derived from prior knowledge.
DPL comprises of a supervision module over K and a prediction module over all input-output pairs (Figure~\ref{fig:DPL}), and defines a probability distribution:
\vspace{-5pt}
\[P(K,Y|X)\propto \prod_v~\Phi_{v}(X, Y)\cdot\prod_i~\Psi(X_i, Y_i)\]

\vspace{-8pt}
Without loss of generality, we assume that virtual evidences are log-linear factors, which can be compactly represented by weighted first-order logical formulas \cite{richardson&domingos06}. Namely, $\Phi_v(X,Y)=\exp(w_v\cdot f_v(X,Y))$, where $f_v(X,Y)$ is a binary feature represented by a first-order logical formula. A hard constraint is the special case when $w_v=\infty$ (in practice, it suffices to set it to a large number, e.g., 10).
In prior use of virtual evidence, $w_v$'s are generally pre-determined from prior knowledge. However, this may be suboptimal. Therefore, we consider a general Bayesian learning setting where each $w_v$ is drawn from a pre-specified prior distribution $w_v\sim P(w_v|\alpha_v)$. Fixed $w_v$ amounts to the special case when the prior is concentrated on the preset value. For uncertain $w_v$'s, we can compute their maximum a posteriori (MAP) estimates and/or quantify the uncertainty.
%Virtual evidences can be compactly specified using probabilistic logic such as Markov logic \cite{richardson&domingos06}, where each formula defines a class of virtual evidences with tied weight. 

\smallpar{Distant supervision}
Virtual evidence for distant supervision is similar to that for direct supervision. For example, for relation extraction, distant supervision from a knowledge base of known relations will set $f_{KB}(X_i,Y_i)=\mathbb{I}[\text{\tt In-KB}(X_i,r) \land Y_i=r]$, where $\text{\tt In-KB}(X_i,r)$ is true iff the entity tuple in $X_i$ is known to have relation $r$ in the KB.
%This heuristics is obviously noisy: the entities may co-occur by chance or for other reasons.

\smallpar{Data programming}
Virtual evidence for data programming is similar to that for distant supervision:
$f_{L}(X_i,Y_i)=\mathbb{I}[L(X_i) = Y_i]$, where $L(X_i)$ is a labeling function provided by domain experts.
Labeling functions are usually high-precision rules, but errors are still common, and different functions may assign conflicting labels to an instance.
Existing denoising strategy assumes that each function makes random errors independently, and resolves the conflicts by weighted votes \cite{ratner&al16}.
In DPL, this can be done by simply treating error probabilities as uncertain parameters and inferring them during learning. 
%At the first-order approximation, the weight for a rule corresponds to the log-odd of its accuracy, and the probability of a label value correlates with the weighted sum of virtual evidences applicable to it. 
%(In general, there could be other factors that influence the weights and label probabilities, which are all accounted for in the graphical model.)

\smallpar{Joint inference}
Constraints on instances or model expectations can be imposed by introducing the corresponding virtual evidence \cite{ganchev&al10} (Proposition 2.1). The weights can be set heuristically \cite{chang&al07,mann&mccallum08,poon&domingos08} or iteratively via primal-dual methods \cite{ganchev&al10}.
In addition to instance-level constraints, DPL can incorporate arbitrary high-order soft and hard constraints that capture the interdependencies among multiple instances.
For example, identical mentions in proximity probably refer to the same entity, which is useful for resolving ambiguous mentions by leveraging their unambiguous coreferences (e.g., an acronym in apposition of the full name).
This can be represented by the virtual evidence $f_{\tt Joint}(X_i,Y_i,X_j,Y_j)=\mathbb{I}[{\tt Coref}(X_i,X_j) \land Y_i=Y_j]$, where ${\tt Coref}(X_i,X_j)$ is true iff $X_i$ and $X_j$ are coreferences.
Similarly, the common denoising strategy for distant supervision replaces the mention-level constraints with type-level constraints \cite{MultiR}. %,lin&al16}.
Suppose that $X_E\subset X$ contains all $X_i$'s with co-occurring entity tuple $E$. The new constraints simply impose that, for each $E$ with known relation $r\in KB$, $Y_i=r$ for at least one $X_i\in X_E$. 
This can be represented by a high-order factor on $(X_i,Y_i: X_i\in X_E)$.

\begin{algorithm}[t]
%\halfspacing
\begin{algorithmic}
\caption{DPL Learning}\label{alg:learn}
\State \textbf{Input:} Virtual evidences $K=\Phi_{1:V}$, deep neural network $\Psi$, inputs $X=(X_1,\cdots,X_N)$, unobserved outputs $Y=(Y_1,\cdots,Y_N)$.
\State \textbf{Output:} Learned prediction module $\Psi^*$ 
\State \textbf{Initialize:} $\Phi^0 \sim \text{priors}$, $\Psi^0 \sim \text{uniform}$.
\For{\texttt{$t=1:T$}}
%\vspace{1pt}
\small{
    \begin{nospaceflalign*}
    %\begin{split}
    %\begin{flalign*}
      q^t(Y) \leftarrow &\arg\min_{q}~D_{KL}(\prod_i~q_i(Y_i)~||~&
      %\nonumber 
      \\
             &\prod_v~\Phi^{t-1}_v(X,Y)\cdot\prod_i~\Psi^{t-1}(X_i,Y_i)) &\\
    %\end{split}
      \Phi^t \leftarrow &\arg\min_{\Phi}~D_{KL}(q^t(Y)~||~ \prod_v~\Phi_v(X,Y)) &\\
      \Psi^t \leftarrow & \arg\min_{\Psi}~D_{KL}(q^t(Y)~||~\prod_i~\Psi(X_i,Y_i)) &
    \end{nospaceflalign*}
    %\end{flalign*}
    }
\EndFor
\State \Return $\Psi^*=\Psi^T$.
\end{algorithmic}
\end{algorithm}

\smallpar{Parameter learning}
Learning in DPL maximizes the conditional likelihood of virtual evidences $P(K|X)$. % by summing out latent $Y$.
We can directly optimize this objective by summing out latent $Y$ to compute the gradient and run backpropagation.
%Directly optimizing this objective is difficult, as $K$ renders $Y$ interdependent on each other, but we can solve it using variational EM.
In this paper, however, we opted for a modular approach using variational EM.
See Algorithm~\ref{alg:learn}.

In the E-step, we compute a variational approximation $q(Y)=\prod_i~q_i(Y_i)$ by minimizing its KL divergence with $P(Y|K,X)$, which amounts to computing marginal probabilities $q_i(Y_i)=P(Y_i|K,X)=\sum_{Y_{-i}}~P(Y_i, Y_{-i}|K,X)$, with current parameters $\Phi, \Psi$.
This is a standard probabilistic inference problem. Exact inference is generally intractable, but there are a plethora of approximate inference methods that can efficiently produce an estimate. We use loopy belief propagation \cite{murphy1999loopy} in this paper, by conducting message passing in $P(K,Y|X)$ iteratively. Note that this inference problem is considerably simpler than end-to-end inference with probabilistic logic, since the bulk of the computation is encapsulated by %the deep neural network factors 
$\Psi$.

%{\bf Specifically ...}
%{\bf inference w/ high-order factors.}
Inference with high-order factors of large size can be challenging, but there is a rich body of literature for handling such structured factors in a principled way. 
In particular, in distant supervision denoising, we alter the message passing schedule so that each at-least-one factor will compute messages to its variables jointly by renormalizing their current marginal probabilities with noisy-or \cite{keith2017identifying}, which is essentially a soft version of dual decomposition \cite{caroe1999dual}.

In the M-step, we treat the variational approximation $q_i(Y_i)$ as probabilistic labels, and use them to optimize $\Phi$ and $\Psi$ via standard supervised learning,  %(with weighted labeled examples), 
which is equivalent to minimizing the KL divergence between the probabilistic labels and the conditional likelihood of $Y$ given $X$ under the supervision module ($\Phi$) and prediction module ($\Psi$), respectively.
For the prediction module, this optimization reduces to standard deep learning. 
Likewise, for the supervision module, this optimization reduces to standard parameter learning for log-linear models (i.e., learning all $w_v$'s that are not fixed). Given the probabilistic labels, it is a convex optimization problem with a unique global optimum. Here, we simply use gradient descent, with the partial derivative for $w_v$ being
$\mathbb{E}_{\Phi(Y,X)}~[f_v(X,Y)] - \mathbb{E}_{q(Y)}~[f_v(X,Y)]$.
For a tied weight, the partial derivative will sum over all features that originate from the same template.
The second expectation can be done by simple counting. The first expectation, on the other hand, requires probabilistic inference in the graphical model. But it can be computed using belief propagation, similar to the E-step, except that the messages are limited to factors within the supervision module (i.e., messages from $\Psi$ are not longer included). 
Convergence is usually fast, upon which %Upon convergence, 
the marginal for each $Y_i$ is available, and $\mathbb{E}_{\Phi(Y,X)}~[f_v(X,Y)]$ is simply the fraction of $Y$ that renders $f_v(X,Y)$ to be true.
Again, this parameter learning problem is much simpler than end-to-end learning with probabilistic logic, as it focuses on refining uncertain weights for indirect supervision, rather than learning complex input patterns for label prediction (handled in deep learning).

\begin{figure}
    \centering
    \includegraphics[width=0.95\linewidth]{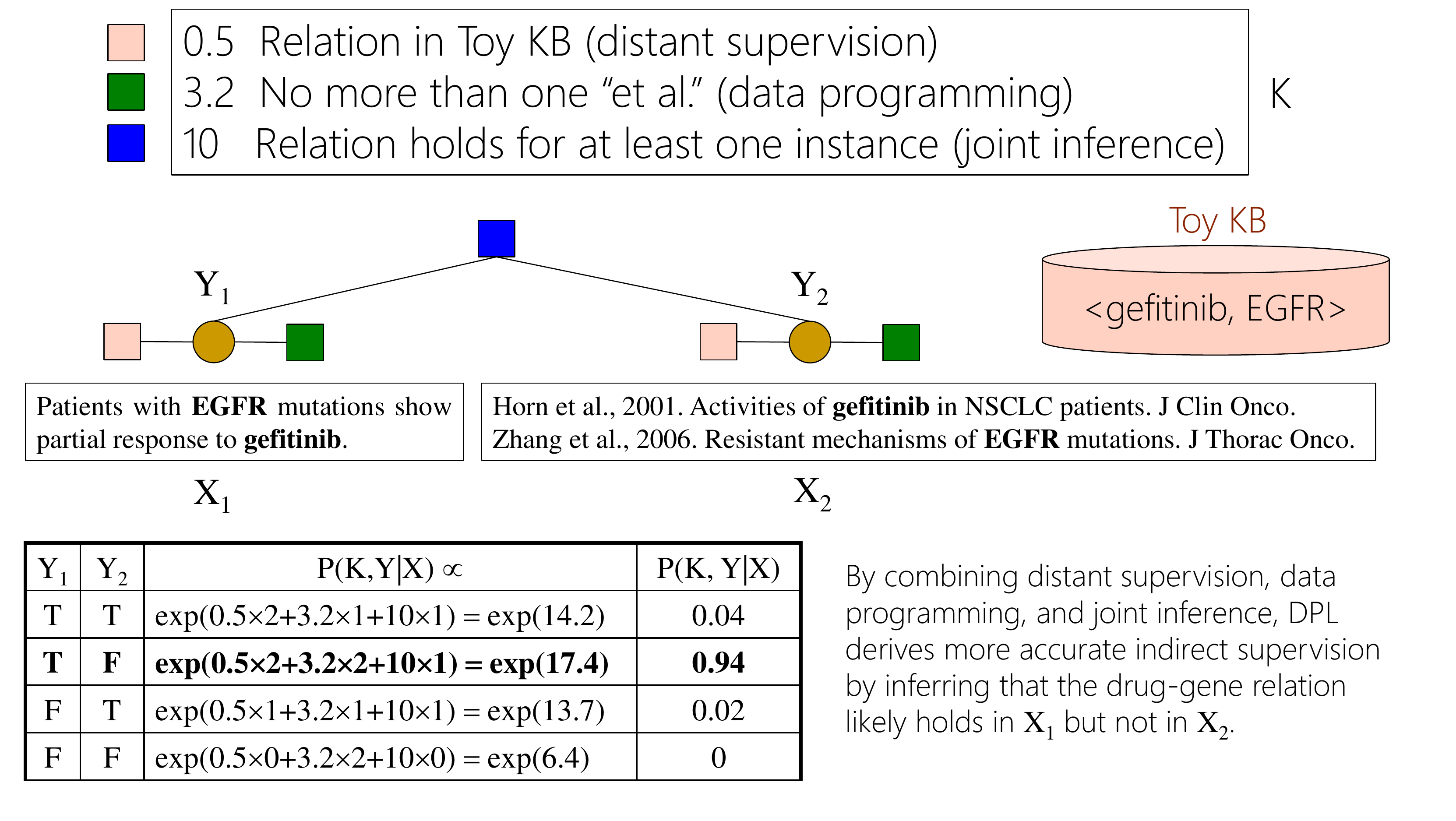}
    \vspace{-10pt}
    \caption{Example of DPL combining various indirect supervision using probabilistic logic. The prediction module is omitted to avoid clutter. 
    }
    \label{fig:dpl-example}
\end{figure}

\smallpar{Example} 
Figure~\ref{fig:dpl-example} shows a toy example on how DPL combines various indirect supervision for predicting drug-gene interaction (e.g., gefitinib can be used to treat tumors with EGFR mutations). Indirect supervision is modeled by probabilistic logic, which defines a joint probability distribution over latent labeling decisions for drug-gene mention pairs in unlabeled text. Here, distant supervision prefers classifying mention pairs of known relations, whereas the data programming formula opposes classifying instances resembling citations, and the joint inference formula ensures that at least one mention pair of a known relation is classified as positive. Formula weight signifies the confidence in the indirect supervision, and can be refined iteratively along with the prediction module. 

\smallpar{Handling label imbalance} One challenge for distant supervision is that negative examples are often much more numerous. A common strategy is to subsample negative examples to attain a balanced dataset. In preliminary experiments, we found that this was often suboptimal, as many informative negative examples were excluded from training. Instead, we restored the balance by up-weighting positive examples.
In DPL, an additional challenge is that the labels are probabilistic and change over iterations. In this paper, we simply used hard EM, with binary labels set using 0.5 as the probability threshold, and the up-weighting coefficient recalculated after each E-step.

\smallsection{Biomedical Machine Reading}

There is a long-standing interest in biomedical machine reading (e.g., \newcite{morgan2008overview, kim2009overview}), but prior studies focused on supervised approaches.
The advent of big biomedical data %such as \$1000-person genomes %heralds the new era of precision medicine \cite{nature}, but 
creates additional urgency for developing scalable approaches that can generalize to new reading tasks. For example, genome sequencing cost has been dropping faster than Moore's Law, yet oncologists can only evaluate tumor sequences for a tiny fraction of patients, due to the bottleneck in assimilating relevant knowledge from publications.
%Machine reading of precision oncology knowledge can be formulated as a relation extraction problem (Figure~\ref{fig:relextract-example}), 
%for which there has not been a labeled corpus available for supervised learning.
Recently, \newcite{peng&al17} formulated precision oncology machine reading as cross-sentence relation extraction (Figure~\ref{fig:relextract-example}) and developed the state-of-the-art system using distant supervision.
While promising, their results still leave much room to improve.
Moreover, they used heuristics to heavily filter entity candidates, with significant recall loss.

In this section, we use cross-sentence relation extraction as a case study for  combining indirect supervision using deep probabilistic logic (DPL).
First, we show that DPL can substantially improve machine reading accuracy in a head-to-head comparison with \newcite{peng&al17}, using the same entity linking method. 
Next, we apply DPL to entity linking itself and attain similar improvement.
Finally, we consider further improving the recall by removing the entity filter. 
%Not surprisingly, this resulted in heavy precision loss using distant supervision alone, as in \newcite{peng&al17}. However, by 
By applying DPL to joint entity linking and relation extraction, we more than doubled the recall in relation extraction while attaining comparable precision as \newcite{peng&al17} with heavy entity filtering.

\smallpar{Evaluation} 
Comparing indirect supervision methods is challenging as there is often
no annotated test set for evaluating precision and recall.
In such cases, we resort to the standard strategy used in prior work by reporting {\em sample precision} (estimated proportion of correct system extractions) and {\em absolute recall} (estimated number of correct system extractions). Absolute recall is proportional to recall and can be used to compare different systems (modulo estimation errors).

\smallpar{Datasets}
We used the same unlabeled text as \newcite{peng&al17}, which consists of about one million full text articles in PubMed Central (PMC)\footnote{\url{www.ncbi.nlm.nih.gov/pmc}}.
Tokenization, part-of-speech tagging, and syntactic parsing were conducted using SPLAT \cite{quirk&al12}, and Stanford dependencies \cite{marneffe&al06} were obtained using Stanford CoreNLP \cite{manning&al14}. 
For entity ontologies, we used DrugBank\footnote{\url{www.drugbank.ca}} and Human Gene Ontology (HUGO)\footnote{\url{www.genenames.org}}. %\footnote{There is no standard ontology for mutations and we resort to using regular expressions.}
DrugBank contains 8257 drugs; we used the subset of 599 cancer drugs.
HUGO contains 37661 genes.
For knowledge bases, we used the Gene Drug Knowledge Database (GDKD)~\cite{dienstmann&al15} and the Clinical Interpretations of Variants In Cancer (CIVIC)\footnote{\url{civic.genome.wustl.edu}}. Together, they contain 231 drug-gene-mutation triples, with 76 drugs, 35 genes and 123 mutations.

%%%%%%%%%%%%%%%%%%%%%%%%%%%%%%%%%%%%%%%%%%%%%%%%%%%%%%%%%%%%%%%%%%%%%%%

\smallsubsection{Cross-sentence relation extraction}

Let $e_1,\cdots,e_m$ be entity mentions in text $T$. 
Relation extraction can be formulated as classifying whether a relation $R$ holds for $e_1,\cdots,e_m$ in $T$.
To enable a head-to-head comparison, we used the same cross-sentence setting as \newcite{peng&al17}, where $T$ spans up to three consecutive sentences and $R$ represents the ternary interaction over drugs, genes, and mutations (whether the drug is relevant for treating tumors with the given gene mutation).

%{\bf There are hundreds of drugs and thousands of genes and mutations. Following Peng et al., we focus on the drug-gene-mutation relation and ignore fine-grained interaction types.}

\smallpar{Entity linking}
In this subsection, we used the entity linker from Literome \cite{poon&al14} to identify drug, gene, and mutation mentions, as in \newcite{peng&al17}. 
This entity linker first identifies candidate mentions by matching entity names or synonyms in domain ontologies, then applies heuristics to filter candidates. The heuristics are designed to enhance precision, at the expense of recall. For example, one heuristics would filter candidates of length less than four, which eliminates key cancer genes such as ER or AKT.
%%%Later, we consider replacing heuristic filtering by applying DPL to train a deep-learning based entity linker with indirect supervision.

\smallpar{Prediction module}
We used the same graph LSTM as in \newcite{peng&al17} to enable head-to-head comparison on indirect supervision strategies.
Briefly, a graph LSTM generalizes a linear-chain LSTM by incorporating arbitrary long-ranged dependencies, such as syntactic dependencies, discourse relations, coreference, and connections between roots of adjacent sentences.
A word might have precedents other than the prior word, and its LSTM unit is expanded to include a forget gate for each precedent. 
%For efficient training, a graph is decomposed into a forward pass and a backward pass, each consisting of edges pointing forward and backward, respectively. Backpropagation is then conducted on these two directed acyclic graphs, similarly to BiLSTM. 
%(If a graph LSTM contains no edges other than word adjacency, it reduces to BiLSTM.) 
See \newcite{peng&al17} for details.

%\textbf{When training, we set different weights for positive and negative instances to make the training is balanced}

\begin{table}[t!]
\footnotesize
\begin{center}
\begin{tabular}{ |p{7.5cm} |}
\hline
\textbf{Distant Supervision}: GDKD, CIVIC\\ 
\hline
%\hline
\textbf{Data Programming (Entity)}\\
%\hline
\text{Mention matches entity name exactly.}\\
\text{Mention not a stop word.}\\
%\textbf{Mention not in a list which contains most common errors made by entity linker.}\\
\text{Mention not following figure designation.}\\
%\textbf{mention's score from entity disambiguation (th=0.55).}\\
%\textbf{Mention's POS tagging contains ``NN".}
\text{Mention's POS tags indicate it is a noun.}\\
\hline
%\hline
\textbf{Data Programming (Relation)}\\ 
%\hline
Less than 30\% of words are numbers in each sentence.\\
No more than three consecutive numbers.\\
No more than two ``et al''.\\
No more than three tokens start with uppercase.\\
No more than three special characters.\\
No more than three keywords indicative of table or figure.\\
Entity mentions do not overlap.\\
\hline 
%\hline
\textbf{Joint Inference}: \text{Relation holds in at least one instance.} \\ 
\hline
\end{tabular}
\vspace{-5pt}
\caption {DPL combines three indirect supervision strategies for cross-sentence relation extraction}
\label{tb:re_factor}	 
\end{center}
\end{table}
%\vspace*{-0.8\baselineskip}

%%%% TODO?
\eat{
+:$\mathbb{I}[\tt \# token \in \{"result", "confer", \ldots \, "response" \} \geq 4]$; 
-:$\mathbb{I}[\tt \# token \in \{"no", "doesn't", \ldots \, "None" \} \le 2]$; 
-:$\mathbb{I}[\tt \# token \in \{"Oncol", "Nat", \ldots \, "PLoS" \} \le 2]$;
-:$\mathbb{I}[{\tt Acronym(D_m, text\_span)} \land D_m={\tt true}, \text{apply the corresponding heuristics to text\_span}]$;
-:$\mathbb{I}[{\tt Acronym(G_m, text\_span)} \land G_m={\tt true}, \text{apply the corresponding heuristics to text\_span}]$;
-:$\mathbb{I}[{\tt Acronym(V_m, text\_span)} \land V_m={\tt true}, \text{apply the corresponding heuristics to text\_span}]$;
-:$\mathbb{I}[{\tt V_m\ is\ a\ variant\ of\ G_m]}$; \\ 
}

%Jnt: EL > 0.6

\smallpar{Supervision module}
We used DPL to combine three indirect supervision strategies for cross-sentence relation extraction (Table~\ref{tb:re_factor}).
For distant supervision, we used GDKD and CIVIC as in \newcite{peng&al17}.
%First, like \newcite{peng&al17}, we used the Gene Drug Knowledge Database (GDKD)~\cite{dienstmann&al15} and the Clinical Interpretations of Variants In Cancer (CIVIC) knowledge base\footnote{\url{http://civic.genome.wustl.edu}} for distant supervision. 
%{\bf Why noisy / Head to head w/ Peng}
%Since these labels are noisy, we used a soft weight $w_{DS}$ for the virtual evidence, with standard $L_2$ prior and initialized as $w^*_{DS}$, a hyperparameter.
For data programming, we introduced labeling functions that aim to correct entity and relation errors.
%based on prior knowledge about this domain, 
%each with its corresponding soft weight, similar to that in distant supervision, with the initial value chosen to reflect our prior belief of the labeling function's accuracy. 
%Some labeling functions focus on preventing entity errors, whereas others on relation ones.
Finally, we incorporated joint inference among all co-occurring instances of an entity tuple with the known relation by imposing the at-least-one constraint (i.e., the relation holds for at least one of the instances).
For development, we sampled 250 positive extractions from DPL using only distant supervision \cite{peng&al17} and excluded them from future training and evaluation.

\eat{
Maintaining a balanced training set (comparable number in positive and negative examples) is generally a good practice. 
Prior weak supervision methods typically ensure this by sampling a comparable number of negative examples after selecting positive ones. {\bf training instances}
}

\smallpar{Experiment results}
%{\bf In practice, convergence happens quickly, usually in 3-5 iterations. Likewise, variational EM overall takes only a few iterations to converge. In our experiments, relation-extraction training took under 30 minutes on a small cluster of 10 nodes.}
We compared DPL with the state-of-the-art system of \newcite{peng&al17}. % on cross-sentence relation extraction.
We also conducted ablation study to evaluate the impact of indirect-supervision strategies.
%Following Peng et al.~\shortcite{peng&al17}, we reported sample precision and absolute recall in PubMed-scale extraction.
For a fair comparison, we used the same probability threshold in all cases (an instance is classified as positive if the normalized probability score is at least 0.5).
For each system, sample precision was estimated by sampling 100 positive extractions and manually determining the proportion of correct extractions by an author knowledgeable about this domain.
Absolute recall is estimated by multiplying sample precision with the number of positive extractions.

\begin{table}[t]
\begin{tabular}{ | l | c | c | c |}
\hline
System & Prec. & Abs. Rec.  & Unique \\ \hline
Peng 2017  & 0.64 & 6768 & 2738 \\ \hline
DPL + $\tt EMB$ & {\bf 0.74} & {\bf 8478} &  {\bf 4821} \\ \hline
DPL & 0.73 & 7666 &  4144 \\ \hline
~~~$-$ $\tt DS$ & 0.29 & 7555 & 4912  \\ \hline
~~~$-$ $\tt DP$ & 0.67 & 4826 &  2629 \\ \hline
~~~$-$ $\tt DP$ $\tt (ENTITY)$ & 0.70 & 7638 &  4074 \\ \hline
~~~$-$ $\tt JI$ & 0.72 & 7418 & 4011  \\ \hline
\end{tabular}
\vspace{-5pt}
\caption {
Comparison of sample precision and absolute recall (all instances and unique entity tuples) in test extraction on PMC.
DPL + $\tt EMB$ is our full system using PubMed-trained word embedding, whereas DPL uses the original Wikipedia-trained word embedding in \newcite{peng&al17}. Ablation: DS (distant supervision), DP (data programming), JI (joint inference).
}
\label{tbl:RE-result}
\end{table}
%\vspace*{-0.8\baselineskip}

\begin{table}[t]
\begin{center}
\begin{tabular}{ | l | c | c | c |}
\hline
Pred. Mod. & Prec. & \ Abs. Rec. & Unique \\ \hline
BiLSTM  & 0.60 & 6243 &  3427 \\ \hline
Graph LSTM  & 0.73 & 7666 &  4144 \\ \hline
\end{tabular}
\vspace{-5pt}
\caption {Comparison of sample precision and absolute recall (all instances and unique entity tuples) in test extraction on PMC. Both use same indirect supervision and Wikipedia-trained word embedding.}
\label{tab:RE_model}
\end{center}
\end{table}

Table~\ref{tbl:RE-result} shows the results. DPL substantially outperformed \newcite{peng&al17}, improving sample precision by ten absolute points and raising absolute recall by 25\%.
Combining disparate indirect supervision strategies is key to this performance gain, as evident from the ablation results. While distant supervision remained the most potent source of indirect supervision, data programming and joint inference each contributed significantly.
Replacing out-of-domain (Wikipedia) word embedding %used in \newcite{peng&al17} 
with in-domain (PubMed) word embedding~\cite{pyysalo&al13} also led to a small gain.

%\vspace*{-0.8\baselineskip}

\newcite{peng&al17} only compared graph LSTM and linear-chain LSTM in automatic evaluation, where distant-supervision labels were treated as ground truth. They found significant but relatively small gains by graph LSTM. We conducted additional manual evaluation comparing the two in DPL. %the same DPL supervision module. 
Surprisingly, we found rather large performance difference, with graph LSTM outperforming linear-chain LSTM by 13 absolute points in precision and raising absolute recall by over 20\% (Table~~\ref{tab:RE_model}). 
This suggests that \newcite{peng&al17} might have underestimated the performance gain by graph LSTM using automatic evaluation.
%We leave further investigation to future work.

\begin{table}[t]
\footnotesize 
\begin{center}
\begin{tabular}{ | p{7.3cm} |}
\hline
\textbf{Distant Supervision}: HGNC\\
\hline %\hline
\textbf{Data Programming}\\
No verbs in POS tags.\\
Mention not a common word. \\
Mention contains more than two characters or one word.\\
More than 30\% of characters are upper case. \\
Mention contains both upper and lower case characters. \\
Mention contains both character and digit. \\
Mention contains more than six characters. \\
Dependency label from mention to parent indicative of direct object.\\
%Mention contains more than two tokens \\
%\textbf{entropy in mention's nearby context.} \\
%Mention 's occur probability calculated in whole dataset. (th=0.02)} \\

\hline %\hline 
\textbf{Joint Inference}\\
Identical mentions nearby probably refer to the same entity.\\
Appositive mentions probably refer to the same entity.\\
Nearby mentions that match synonyms of same entity probably refer to the given entity.
\\ \hline 
\end{tabular}
\vspace{-5pt}
\caption {DPL combines three indirect supervision strategies for entity linking.}
\label{tb:EL-VE}	 
\end{center}
\end{table}
\eat{
-:$\mathbb{I}[{\tt strcase(m)} \land G_m={\tt true}]$; 
-:$\mathbb{I}[{\tt token\_case(m)} \land G_m={\tt true}]$; 
-:$\mathbb{I}[{\tt context\_entropy(m)} \land G_m={\tt true}]$;
-:$\mathbb{I}[{\tt entropy(G_{m})} \land G_m={\tt true}]$; 
-:$\mathbb{I}[{\tt special\_char(m)} \land G_m={\tt true}]$; 
-:$\mathbb{I}[{\tt Acronym(m, text\_span)} \land G_m={\tt true}, \text{apply the heuristics to text\_span}]$  
}

%%%%%%%%%%%%%%%%%%%%%%%%%%%%%%%%%%%%%%%%%%%%%%%%%%%%%%%%%%%%%%%%%%%%%%%%%%%%%%%

\smallsubsection{Entity linking}

Let $m$ be a mention in text and $e$ be an entity in an ontology. The goal of entity linking is to predict $\tt Link(m,e)$, which is true iff $m$ refers to $e$, for every candidate mention-entity pair $m,e$.
We focus on genes in this paper, as they are particularly noisy.
%This is a prerequisite step for relation extraction. 
%Figure~\ref{fig:EL-example} shows an example, where ``Peptide Deformylase, Mitochondrial'' and ``PDF'' both refer to the gene $\tt PDF$ (HGNC:30012).

%\smallpar{Unlabeled text}
%Genes, drugs, diseases are some of the entity types most pertinent to precision medicine. 
%We focus on genes, drugs, and mutations in cross-sentence relation extraction.
%We used the human gene ontology HGNC \footnote{\url{www.genenames.org}}.
%For drugs, we used DrugBank (\url{www.drugbank.ca}).
%For mutations, we followed Peng et al.~\shortcite{peng&al17} and focused on point mutations, with the standard form $\tt [A-Za-z][1-9]+[A-Za-z]$ (representing amino acid change at certain offset).
%We used a more recent version of PMC in early 2017, with 1.5 million articles.

\smallpar{Prediction module} We used BiLSTM with attention over the ten-word windows before and after a mention.
The embedding layer is initialized by word2vec embedding trained on PubMed abstracts and full text~\cite{pyysalo&al13}.
%, and a multi-layer perceptron is used at the top for classification.
The word embedding dimension was 200. We used 5 epochs for training, with Adam as the optimizer. We set learning rate to 0.001, and batch size to 64. 
%\textbf{When training, we set different weights for positive and negative instances to make the training is balanced} 

\smallpar{Supervision module}
As in relation extraction, we combined three indirect supervision strategies using DPL (Table~\ref{tb:EL-VE}).
For distant supervision, we obtained all mention-gene candidates by matching PMC text against the HUGO lexicon. We then sampled a subset of 200,000 candidate instances as positive examples. We sampled a similar number of noun phrases as negative examples. 
For data programming, we introduced labeling functions that used mention characteristics (longer names are less ambiguous) or syntactic context (genes are more likely to be direct objects and nouns). 
For joint inference, we leverage linguistic phenomena related to coreference (identical, appositive, or synonymous mentions nearby are likely coreferent).
%For development, we randomly sampled 100 positive extractions by DPL using only distant supervision and excluded them from training and evaluation.
\eat{
\begin{table}[t]
	\begin{center}
		\begin{tabular}{ | l | c | c | c | c| }
			\hline
			System & Acc. & F1 & Prec. & Rec. \\ \hline
			String Match & 0.18 & 0.31 & 0.18 & 1.00 \\ \hline
			DS & 0.64 & 0.43 & 0.28 & 0.90 \\ \hline
			DS + DP   & 0.66 & 0.68 & 0.74 & 0.62\\ \hline
			DS + DP + JI & {\bf 0.73} & {\bf 0.76}   & 0.70 & 0.84   \\ \hline
		\end{tabular}
		\vspace{-5pt}
\caption {Comparison of gene entity linking results on a balanced test set. The string-matching baseline has low precision. By combining indirect supervision strategies, DPL substantially improved precision while retaining reasonably high recall.
}
		\label{tab:EL}	 
	\end{center}
\end{table}
}

\begin{table}[t]
	\begin{center}
		\begin{tabular}{ | l | c | c | c | c| }
			\hline
			System & Acc. & F1 & Prec. & Rec. \\ \hline
			String Match & 0.18 & 0.31 & 0.18 & 1.00 \\ \hline
			DS & 0.64 & 0.71 & 0.62 & 0.83 \\ \hline
			DS + DP   & 0.66 & 0.71 & 0.62 & 0.83\\ \hline
			DS + DP + JI & {\bf 0.70} & {\bf 0.76}   & 0.68 & 0.86   \\ \hline
		\end{tabular}
		\vspace{-5pt}
\caption {Comparison of gene entity linking results on a balanced test set. The string-matching baseline has low precision. By combining indirect supervision strategies, DPL substantially improved precision while retaining reasonably high recall.
}
		\label{tab:EL}	 
	\end{center}
\end{table}

%\vspace*{-0.8\baselineskip}

\begin{table}[t]
\begin{center}
\begin{tabular}{ | l | c | c | c |}
\hline
 & F1 & Precision & Recall \\ \hline
GNormPlus  & 0.78 & 0.74 & 0.81 \\ \hline
DPL  & 0.74 & 0.68 & 0.80 \\ \hline
\end{tabular}
\vspace{-5pt}
\caption {Comparison of gene entity linking results on BioCreative II test set. GNormPlus is the state-of-the-art system trained on thousands of labeled examples. DPL used only indirect supervision.}
\label{tab:GNormPlus}
\end{center}
\end{table}

\smallpar{Experiment results}
For evaluation, we annotated a larger set of sample gene-mention candidates and then subsampled a balanced test set of 550 instances (half are true gene mentions, half not).
%randomly sampled and annotated a balanced test set of 200 instances of gene-mention candidates (half are really gene mentions, half not). 
These instances were excluded from training and development.
Table~\ref{tab:EL} compares system performance on this test set.
The string-matching baseline has a very low precision, as gene mentions are highly ambiguous, which explains why \newcite{peng&al17} resorted to heavy filtering.
By combining indirect supervision strategies, DPL improved precision by over 50 absolute points, while retaining a reasonably high recall (86\%). All indirect supervision strategies contributed significantly, as the ablation tests show.
We also evaluated DPL on BioCreative II, a shared task on gene entity linking  \cite{morgan2008overview}. 
We compared DPL with GNormPlus \cite{wei2015gnormplus}, the state-of-the-art supervised system trained on thousands of labeled examples in BioCreative II training set.
Despite using zero manually labeled examples, DPL attained comparable F1 and recall (Table~\ref{tab:GNormPlus}). The difference is mainly in precision, which indicates opportunities for more indirect supervision.

\smallsubsection{Joint entity and relation extraction}

%need to say something here, \cite{ren2017cotype} did something similar, but we're slightly different since we didn't really do joint: we did a pipiline level, but the RE result doesn't flow to EL, only EL to RE.

An important use case for machine reading is to improve knowledge curation efficiency by offering extraction results as candidates for curators to vet. The key to practical adoption is attaining high recall with reasonable precision~\cite{peng&al17}.
The entity filter used in \newcite{peng&al17} is not ideal in this aspect, as it substantially reduced recall. 
In this subsection, we consider replacing the entity filter by the DPL entity linker Table~\ref{tab:joint}. % (prior subsection). % in relation-extraction training and prediction.
Specifically, we added one labeling function to check if the entity linker returns a normalized probability score above $p_{\tt TRN}$ for gene mentions, and filtered test instances if the gene mention score is lower than $p_{\tt TST}$. 
%, with $p_{\tt TRN}$ and $p_{\tt TST}$ being hyperparameters. 
We set $p_{\tt TRN}=0.6$ and $p_{\tt TST}=0.3$ from preliminary experiments. The labeling function discouraged learning from noisy mentions, and the test-time filter skips an instance if the gene is likely wrong. 
%Table~\ref{tab:joint} shows the results. 
Not surprisingly, without entity filtering, \newcite{peng&al17} suffered large precision loss.
All DPL versions substantially improved accuracy, with significantly more gains using the DPL entity linker.

\begin{table}[t]
\begin{center}
\begin{tabular}{ | l | c | c | c |}
\hline
System & Prec & Abs. Rec. & Unique \\ \hline
Peng 2017 &	0.31 & 11481 & 5447 \\ \hline
DPL (RE)  & 0.52 & 17891 & 8534 \\ \hline
~$+$ EL (TRN) & 0.55 & {\bf 21881} &  {\bf 11047} \\ \hline
~$+$ EL (TRN/TST) & {\bf 0.61} & 20378 &  10291 \\ \hline
\end{tabular}
\vspace{-5pt}
\caption{Comparison of sample precision and absolute recall (all instances and unique entity tuples) when all gene mention candidates are considered. \newcite{peng&al17} used distant supervision only. RE: DPL relation extraction. EL: using DPL entity linking in RE training (TRN) and/or test (TST).
}
\label{tab:joint}	 
\end{center}
\end{table}

\begin{table}[t]
\begin{center}
\begin{tabular}{ | c | c | c| c | c|}
\hline
Gene & Drug & Mut. & Gene-Mut. & Relation\\
\hline
27\% & 4\% & 20\% & 45\% & 24\% \\ \hline
\end{tabular}
\vspace{-5pt}
\caption {Error analysis for DPL relation extraction.}
\label{tab:error_analyis}	 
\end{center}
\end{table}

\smallsubsection{Discussion}

\smallpar{Scalability} 
%DPL was partially motivated by the computational challenges in end-to-end modeling with probabilistic logic. DPL mitigates this problem by limiting probabilistic logic to modeling instance-level labels. Text-level modeling (words, n-grams, etc.) is relegated to deep learning (graph LSTM in this paper). In relation extraction, each instance contains in average 70 words, so if we were to include text-level modeling, the number of random variables would increase by at least 70 folds, even if we restricted to bag-of-word features. More importantly, deep learning implicitly conducts feature learning by distilling lexical and long-ranged textual patterns, effectively exploring an exponentially large feature space via continuous optimization. This would be intractable in probabilistic logic and would require heavy approximation like greedy search. 
DPL is efficient to train, taking around 3.5 hours for relation extraction and 2.5 hours for entity linking in our PubMed-scale experiments, with 25 CPU cores (for probabilistic logic) and one GPU (for LSTM). For relation extraction, the graphical model of probabilistic logic contains around 7,000 variables and 70,000 factors. At test time, it is just an LSTM, which predicted each instance in less than a second. 
In general, DPL learning scales linearly in the number of training instances. For distant supervision and data programming, DPL scales linearly in the number of known facts and labeling functions. As discussed in Section 3, joint inference with high-order factors is more challenging, but can be efficiently approximated. For inference in probabilistic logic, we found that loopy belief propagation worked reasonably well, converging after 2-4 iterations. Overall, we ran variational EM for three iterations, using ten epochs of deep learning in each M-step. We found these worked well in preliminary experiments and used the same setting in all final experiments.

\begin{figure}
    \centering
    \includegraphics[width=0.95\linewidth]{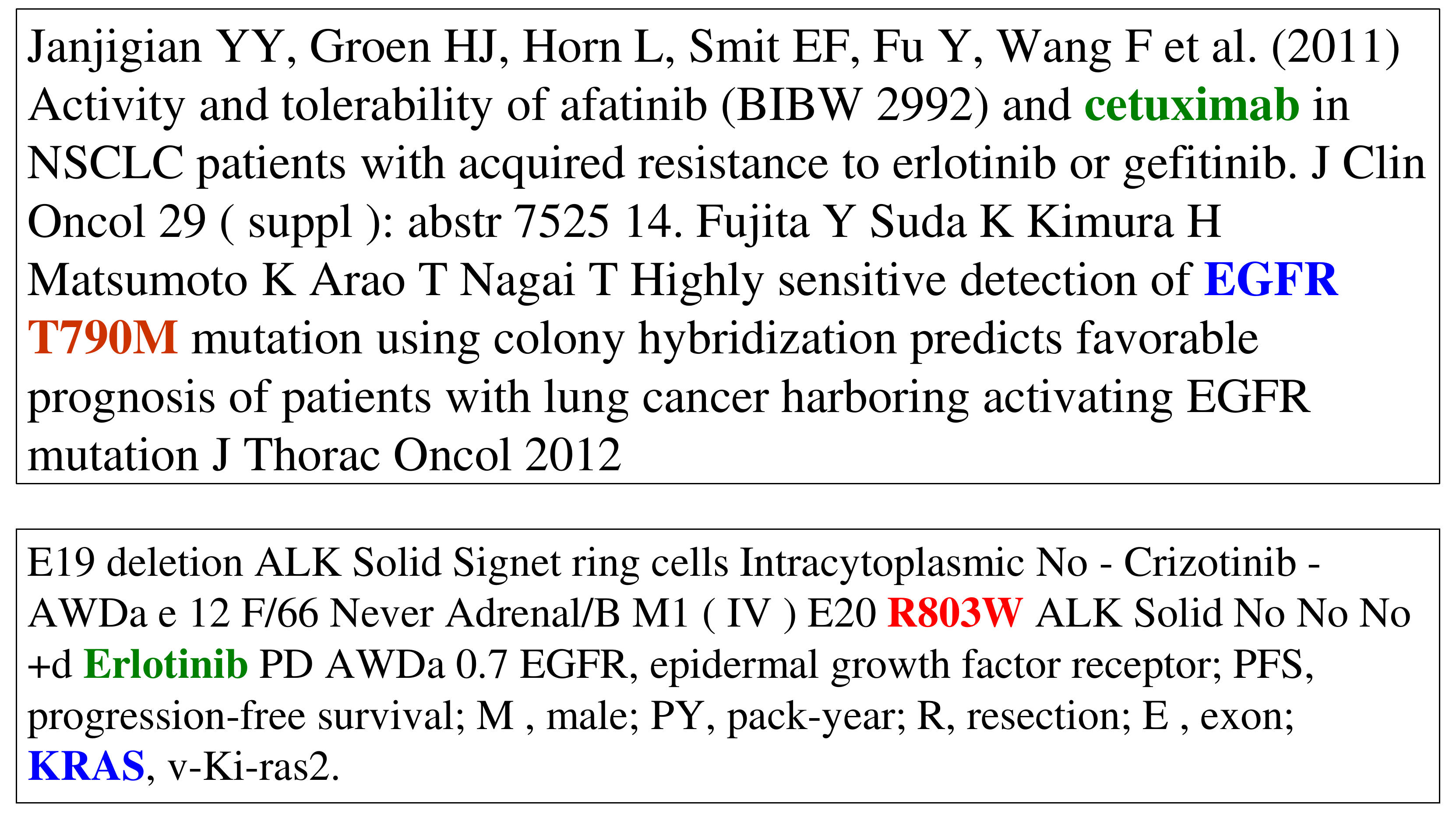}
    \vspace{-5pt}
    \caption{Example of relation-extraction errors corrected by DPL with additional indirect supervision. 
    }
    \label{fig:fix-example}
\end{figure}

\smallpar{Accuracy}
To understand more about DPL's performance gain over distant supervision, we manually inspected some relation-extraction errors fixed by DPL after training with additional indirect supervision. 
Figure~\ref{fig:fix-example} shows two such examples. 
While some data programming functions were introduced to prevent errors stemming from citations or flattened tables, none were directly applicable to these examples. This shows that DPL can generalize beyond the original indirect supervision.

While the results are promising, there is still much to improve. 
Table~\ref{tab:error_analyis} shows estimated precision errors for relation extraction by DPL. (Some instances have multiple errors.) 
%Entity linking is the clearly the key for further improvement. 
Entity linking can incorporate more indirect supervision.
Joint entity linking and relation extraction can be improved by feeding back extraction results to linking. 
Improvement is also sorely needed in classifying mutations and gene-mutation associations.
The prediction module can also be improved, e.g., by adding attention to graph LSTM.
%and recognizing other relevant entities in same text span for better predictions. 
DPL offers a flexible framework for exploring all these directions.

\smallsection{Conclusion}

We introduce DPL as a unifying framework for indirect supervision, by composing probabilistic logic with deep learning. Experiments on biomedical machine reading show that this enables novel combination of disparate indirect supervision methodologies, resulting in substantial gain in accuracy. Future directions include: combining DPL with deep generative models; exploring alternative optimization strategies; applications to other domains. 

\smallsection{Acknowledgements}
We thank David McAllester, Chris Quirk, and Scott Yih for useful discussions, and the three anonymous reviewers for helpful comments.

\bibliographystyle{acl_natbib_nourl}
\bibliography{ref_noise}

%\bibliographystyle{acl_natbib}
%\bibliography{ref_noise}

\end{document}